\pdfoutput=1

\documentclass[11pt]{article}

\usepackage[final]{acl}

\usepackage{times}
\usepackage{latexsym}

\usepackage[T1]{fontenc}

\usepackage[utf8]{inputenc}

\usepackage{microtype}

\usepackage{inconsolata}

\usepackage{hyperref}       
\usepackage{url}            
\usepackage{booktabs}       
\usepackage{amsfonts}       
\usepackage{nicefrac}       
\usepackage{microtype}      
\usepackage{xcolor}        
\usepackage{times}
\usepackage{epsfig}
\usepackage{graphicx}
\usepackage{dsfont}
\usepackage{multirow}
\usepackage{enumerate}
\usepackage{enumitem}
\usepackage{pifont}
\usepackage{comment}
\usepackage{caption}
\usepackage{silence}
\usepackage{physics}
\usepackage{nicematrix}
\usepackage{wrapfig}
\usepackage{float}
\usepackage{tikz}
\usepackage{bm}
\usepackage{bbm}
\usepackage{makecell}
\usepackage{soul}

\usepackage{colortbl}
\definecolor{mygray}{gray}{0.9}

\newcommand{\cmark}{\ding{52}}%
\newcommand{\fmark}{\ding{56}}%
\def\halfcheckmark{\textcolor{black}{\ding{52}}{\small\textcolor{black}{\kern-0.7em\ding{55}}}}

\newcommand{\ie}{i.e.}
\newcommand{\eg}{e.g.}

\title{Beyond Literal Descriptions: Understanding and Locating Open-World Objects Aligned with Human Intentions}

\author{
  Wenxuan Wang\textsuperscript{1,2,3}\thanks{Equal contribution.}, 
  \textbf{Yisi Zhang}\textsuperscript{4}\footnotemark[1]~,
  \textbf{Xingjian He}\textsuperscript{1},
  \textbf{Yichen Yan}\textsuperscript{1,2},
  \textbf{Zijia Zhao}\textsuperscript{1,2},\\
  \textbf{Xinlong Wang}\textsuperscript{3},
  \textbf{Jing Liu}\textsuperscript{1,2}\thanks{Corresponding author.}\\
  \textsuperscript{1}Institute of Automation, Chinese Academy of Sciences (CASIA)\\
  \textsuperscript{2}School of Artificial Intelligence, University of Chinese Academy of Sciences (UCAS) \\
  \textsuperscript{3}Beijing Academy of Artificial Intelligence (BAAI)\\
  \textsuperscript{4}University of Science and Technology Beijing (USTB)\\
  \texttt{wangwenxuan2023@ia.ac.cn}, \texttt{wangxinlong@baai.ac.cn}, \texttt{jliu@nlpr.ia.ac.cn} \\
}

\begin{document}
\maketitle

\begin{abstract}

Visual grounding (VG) aims at locating the foreground entities that match the given natural language expressions.
Previous datasets and methods for classic VG task mainly rely on the prior assumption that the given expression must literally refer to the target object, 
which greatly impedes the practical deployment of agents in real-world scenarios.
Since users usually prefer to provide intention-based expression for the desired object instead of covering all the details, it is necessary for the agents to interpret the intention-driven instructions. 
Thus, in this work, we take a step further to the intention-driven visual-language (V-L) understanding.
To promote
classic VG towards human intention interpretation, we propose a new intention-driven visual grounding (IVG) task and build a large-scale IVG dataset termed IntentionVG with free-form intention expressions.
Considering that practical agents need to move and find specific targets among various scenarios to realize the grounding task, 
our IVG task and IntentionVG dataset have taken the crucial properties of both multi-scenario perception and egocentric view into consideration.
Besides, various types of models are set up as the baselines to realize our IVG task. 
Extensive experiments on our IntentionVG dataset and baselines demonstrate the necessity and efficacy of our method for the V-L field. 
To foster future research in this direction, our newly built dataset and baselines will be publicly available at \url{https://github.com/Rubics-Xuan/IVG}.

\end{abstract}

\section{Introduction}
\label{introduction}

Recently, the research community has witnessed the rapid advancement of multimodal embodied intelligence \cite{ahn2022can,reed2022generalist,driess2023palm,shah2023lm,gao2023physically,brohan2023rt}. 
For an intelligent agent, the capability of locating the target objects in the unpredictable open-world scenarios based on natural language expressions is crucial, underscoring the importance of visual grounding (VG) task within the broader context. 
Notably, instructions provided by users often encapsulate their genuine needs through nuanced intention-driven expressions, which are usually not literal or explicit with much details as classic VG task. 
This nuance brings to light the critical role of VG based on user intention expressions, where challenge lies in interpreting and responding to user commands in a way that truly reflect their underlying desires, 
transcending surface-level expressions to foster more flexible and understanding human-machine interactions.

\begin{figure*}[htbp]
  \centering
  \includegraphics[width=0.92\linewidth]{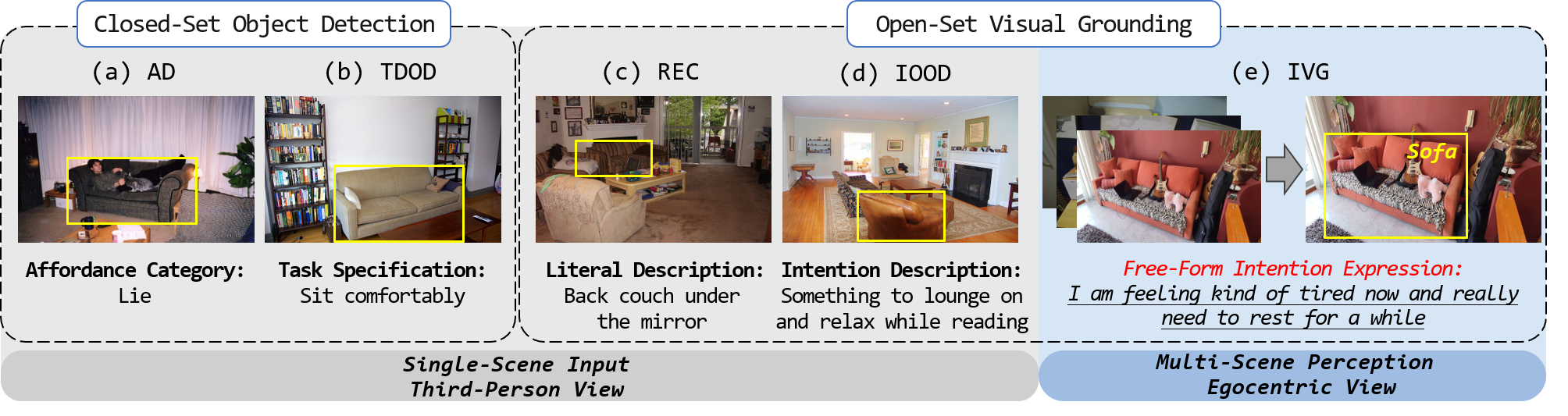}
  \vspace{-8pt}
  \caption{Task Comparison between Affordance Detection (AD), Task-Driven Object Detection (TDOD), Referring Expression Comprehension (REC), Intention-Oriented Object Detection (IOOD) and Intention-driven VG (IVG).}
  \vspace{-5mm}
  \label{fig:intro}
\end{figure*}

Contrary to the classic VG task, realizing intention-based grounding involves several unique aspects that require consideration.
1) \emph{Intention-Driven Descriptions}: Since humans tend to provide intention-based expressions to get the desired objects rather than detailing every aspect, it is imperative for intelligent agents to interpret these intention-driven instructions and act accordingly, focusing on the semantic core of the requests rather than their literal descriptions. 
However, previous studies in this field have primarily concentrated on literal textual descriptions, with scant attention to understanding user intentions.
2) \emph{Egocentric Perspective}: As explored in prior studies \cite{qi2020reverie,kurita2023refego,zhu2023egoobjects,lee2023determinet}, the practical agents actually receive all visual information from a first-person view. 
However, most classic VG datasets are predominantly collected from third-person perspective, which greatly deviate from the application contexts of an embodied agent.
3) \emph{Multi-Scene Perception}: Given that agents are expected to navigate and identify specific targets across diverse scenarios in the real world, the capability to perceive and interact within dynamic multi-scene environments is crucial for accurately accomplishing visual grounding task. Yet, most prior research on VG has overlooked this critical aspect, focusing mainly on static, single-scene visual input.
\ul{In summary, while intention-based VG is highly meaningful for multimodal embodied intelligence, there is a notable scarcity of research on visual grounding based on human intentions, and the current lack of relevant data further compounds the challenge of this task}.

Therefore, in this work, we attempt to fill this important blank space that has been neglected before and move towards intention-oriented visual grounding.
Specifically, based on classic VG, we propose a new intention-driven visual grounding (IVG) task to push towards intention-oriented vision-language understanding, which requires the models to identify the corresponding scene and target object that match the intent expressions from the given multi-scene input.
To solve the data scarcity problem of intention-oriented grounding task, we build a large-scale grounding dataset termed IntentionVG which is also the first grounding dataset to support free-form intention expressions.
Besides, we also construct several baseline models as straightforward solutions to our proposed IVG task, including both zero-shot \& fine-tuning settings and integrated \& end-to-end model types. 
The constructed baselines set the new state-of-the-art (SOTA) performance on our IntentionVG benchmark dataset for IVG task, which leaves further room for achieving performance improvement by future research.

Overall, our main contributions of this work can be summarized as follows:
\setlist{nolistsep}
\begin{itemize}[noitemsep,leftmargin=*]
    \item 
    We propose a new IVG task (as presented in Fig. \ref{fig:intro}) and introduce a new setting based on egocentric viewpoint with multi-scene perception to better evaluate the embodied agents’ perceiving ability, 
    transcending the classic VG task towards better understanding of human desires in the open-world scenarios. 
    \item 
    We build an intention-oriented grounding benchmark named IntentionVG, which to the best of our knowledge is the first large-scale intention-driven grounding dataset  that supports free-form intention-based vision-language annotations.
    \item 
    We develop a series of baseline models under both zero-shot and fine-tuning settings to effectively realize precise vision-language understanding for our IVG task, setting new SOTA performance on our IntentionVG benchmark dataset.
\end{itemize}

\section{Related Work} 
\label{sec:related_work}

\paragraph{Classic Visual Grounding}\!\!\!is to locate the target object corresponding to the given natural language expression in an image. 
The two fundamental VG tasks are distinguished by their output form.
Referring Expression Comprehension (REC) \cite{mao2016generation,chen2018real,deng2021transvg,deng2023transvg++,bai2023qwen,zhu2023minigpt,chen2023minigpt,chen2023shikra,you2023ferret} and Referring Expression Segmentation (RES) \cite{hu2016segmentation,ye2019cross,ding2021vision,yang2022lavt,wang2022cris,lai2023lisa,zou2023segment,zou2023generalized,wang2024cm} have been well studied by previous works, among which REC is the main focus of this work given its heightened significance.
However, previous studies in this field are mainly stuck on the literal description based grounding with a single input image, and the images among previous datasets are typically collected in a third-person perspective. 
In this work, we pioneeringly propose a new IVG task and an intention-based grounding dataset IntentionVG based on egocentric view and multi-scene perception, pushing towards human intention understanding in the practical scenarios.

\begin{figure*}[htbp]
    \centering
    \includegraphics[width=0.65\textwidth]{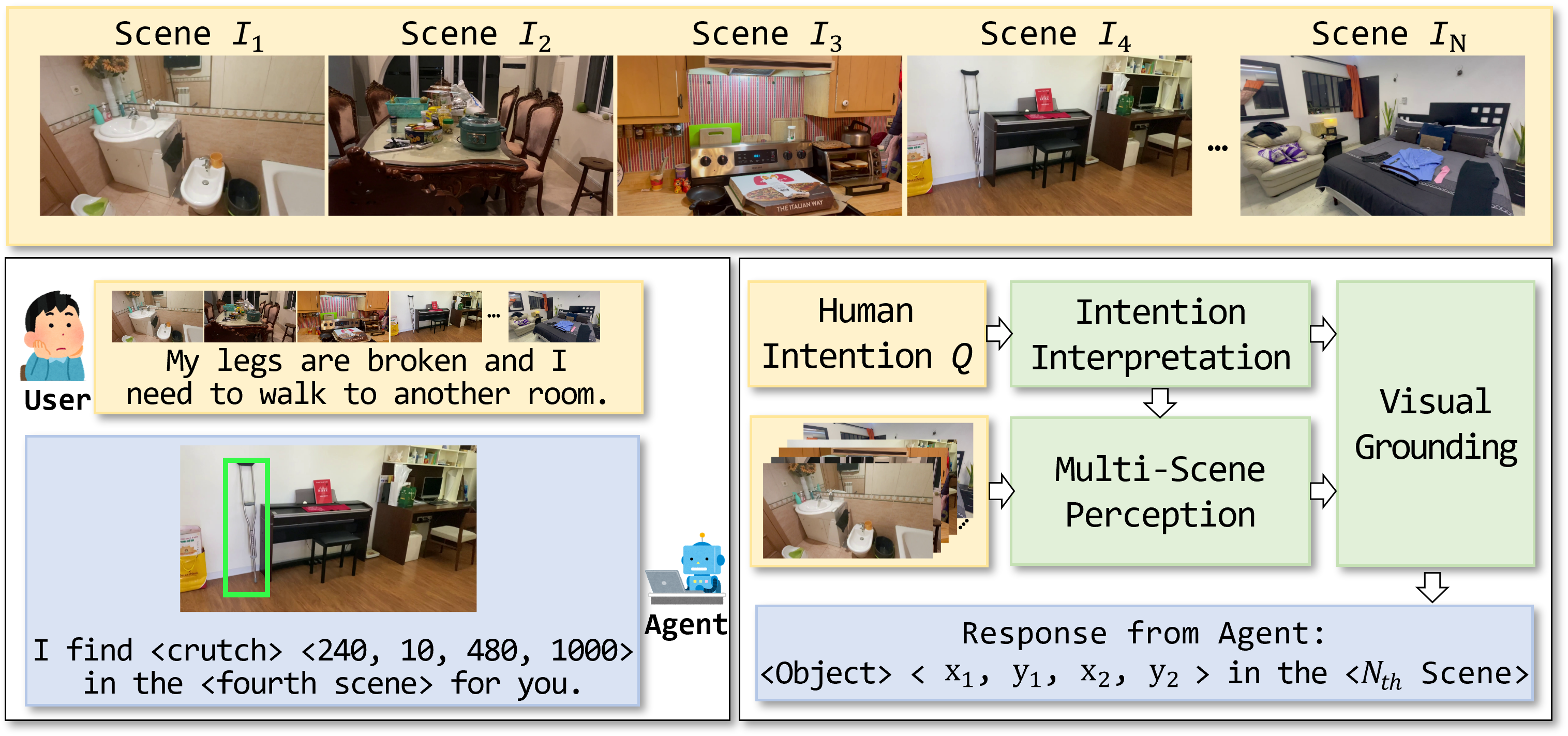}
    \vspace{-5pt}
    \caption{
    The illustration about the overall pipeline of our intention-driven visual grounding task, which mainly comprises intention interpretation, multi-scene perception and the subsequent visual grounding.
    }
    \label{fig_task_definition}
    \vspace{-15pt}
\end{figure*}

\paragraph{Vision-Language Complex Reasoning}\!\!\!aims at understanding intricate textual-visual input information and accomplishing the vision-language (V-L) tasks based on reasoning, in which broader knowledge and strong expression comprehension ability are essential thus posing a greater challenge compared with conventional V-L tasks. 
Due to the reasoning ability and rich prior knowledge in large language models (LLMs), numerous methods such as \cite{pi2023detgpt,zhao2023bubogpt,you2023ferret,li2024lego,chen2024mapgpt} leverage LLMs to understand complex instructions. 
However, prior works mainly focus on literal description reasoning, falling short in understanding potential user intentions. 
Recently, LISA \cite{lai2023lisa} introduces a challenging RES benchmark that incorporates complex expressions, and RIO \cite{qu2023rio} proposes a new IOOD dataset that includes the specific affordance of the objects. 
The format of RIO's sentences is \emph{"Something can be used..."}, which fails to describe the intentions directly from the user's perspective but simply describes the affordance of the target objects. 
Therefore, in this work, starting from a perspective that aligns more closely with real-world scenarios, we for the first time integrate egocentric viewpoint of data collection, multi-scene perception, and free-form expression of human intentions to construct a new IVG task with a corresponding IntentionVG benchmark dataset, thereby enhancing LLMs' reasoning capabilities to better understand human intentions.

\section{IVG Task \& IntentionVG Dataset} 
\label{sec:data}

In this section, we first introduce our IVG task's definition (Sec.~\ref{subsec:task_definition}) and present collection pipeline for IntentionVG data (Sec.~\ref{subsec:data_engine}).
Then, the specific details and evaluation metrics about our IntentionVG are provided (Sec.~\ref{subsec:dataset_details} and Sec.~\ref{subsec:metrics}).

\subsection{IVG Task Description}
\label{subsec:task_definition}

As presented in Fig. \ref{fig_task_definition}, the visual-textual input and corresponding ground truth consist of a human intention query $Q$, a set of scene candidates $I_{1}, ..., I_{N}$, a positive scene index $N_{th}$, and a target bounding box $(x_1,y_1,x_2,y_2)$ together with its object category <Object> in the positive scene image $I_{N_{th}}$. 
The overall pipeline of our proposed IVG task can be decomposed into two stages. 
The first stage (\ie, intention interpretation and multi-scenario perception) is aimed at identifying the target scene image $I_{N_{th}}$ from a predefined set of potential scenes that aligns mostly with the given intention expressions, based upon query $Q$ made by users. 
In this phase, it is imperative for the models to comprehend the textual queries posed by humans in conjunction with the observed visual scenes, and subsequently make correct judgments by returning the correct scene index $N_{th}$. 
The second stage (\ie, visual grounding) involves the localization of specific object within the chosen scene image, returning the target bounding box $(x_1,y_1,x_2,y_2)$ and its category tag <Object>. 
In essence, our new IVG task necessitates the model's proficiency in concurrently understanding both user intention-based requests and multi-scene visual inputs, as well as the models' capability to perform scene selection and visual localization in alignment with the underlying human intentions. 
The complete response can be organized into the following format: 

\vspace{-5mm}
\begin{equation}
    <\!\!Object\!\!>\!\!(x_1,y_1,x_2,y_2) \ in \ <\!\!N_{th} \ Scene\!\!>
\end{equation}
\vspace{-8mm}

\subsection{Data Collection Engine}
\label{subsec:data_engine}

\begin{figure*}[htbp]
  \centering
  \includegraphics[width=0.65\textwidth]{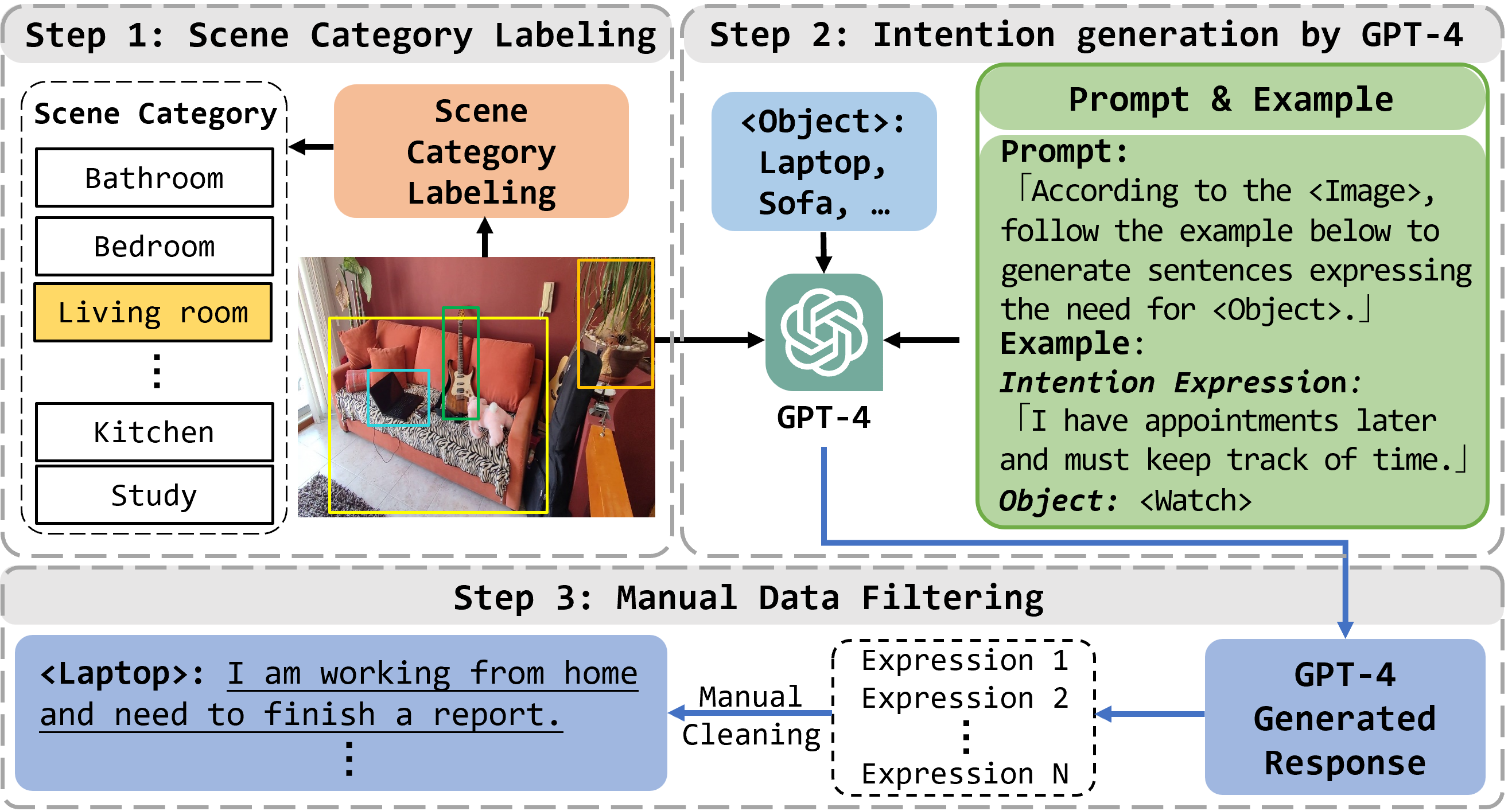}
  \vspace{-5pt}
  \caption{
  The illustration of data collection engine for IntentionVG. 
  We start by inheriting EgoObjects \cite{zhu2023egoobjects} data and conduct scene category labeling for each image. 
  Then we feed GPT-4 with V-L input to generate the draft of intention-driven response.
  At last, we conduct data filtering by manually selecting the well matched expression-bounding box (bbox) pairs.
  }
  \label{fig_data_collection_engine}
  \vspace{-5mm}
\end{figure*}

As shown in Fig. \ref{fig_data_collection_engine}, we build our IntentionVG dataset based on the egocentric grounding dataset EgoObjects \cite{zhu2023egoobjects}, inheriting the annotations of object categories and bounding boxes. 
The entire data collection process involves three steps. 
Since the practical applications require the agents capable of conducting visual perception among diverse scenes, we first conduct scene category labeling of each image.
Specifically, the inherited EgoObjects data is manually annotated with indoor scene categories, resulting in a total of 10 scene classes.
In the second step, with the rapid developments of multimodal large language models, we take advantage of GPT-4 \cite{achiam2023gpt} to adeptly comprehend the visual relationship between object categories and given images, generating their associated intention expressions.
Utilizing the collected images and object category information, we craft well-designed prompts with examples (\ie, \emph{"According to the <Image>, follow the examples below to generate sentences expressing the need for <object>. Example: [I have appointments later and must keep track of time] for <watch> ..."}) to query GPT-4 for the expected outputs. 
At last, we manually review and refine the response generated by GPT-4, making sure that the intention expression of each object is objectively aligned with the scene category of the image. 
Subsequently, based on the proportional distribution within each scene category, we partition our IntentionVG data into training and testing sets with 98,269 and 3,379 images, respectively.

\subsection{IntentionVG Dataset Details}
\label{subsec:dataset_details}

\vspace{-5pt}
\begin{table}[htbp]
    \centering
    \small
    \setlength{\tabcolsep}{0.4pt} 
    \begin{tabular}{lccccc}
    \specialrule{.1em}{.05em}{.05em}
        Datasets & \#Imgs & \#Labels & Intentions & \#Cats & \#Avg Len \\
        \midrule
        \multicolumn{6}{l}{\color{gray}Classic Visual Grounding} \\
        ReferIt & 20K & 97K & -- & 238 & 3.2 \\
        RefCOCO & 20K & 50K & -- & 80 & 3.6 \\
        RefCOCO+ & 20K & 49K & -- & 80 & 3.5 \\
        RefCOCOg & 26K & 54K & -- & 80 & 8.4 \\
        GRES & 20K & 60K & -- & 80 & 3.7 \\
        \midrule
        \multicolumn{6}{l}{\color{gray}Referring Video Object Segmentation} \\
        Refer-Youtube-VOS & 4K & 7K & - & 94 & Unknown \\
        RefEgo  & 12K & 12K & - & 505 & 13.4 \\
        \midrule
        \multicolumn{6}{l}{\color{gray}Affordance Detection} \\
        ADE-Aff & 10K & 26K & Verbs & 150 & / \\
        PAD  & 4K & 4K & Verbs & 72 & / \\
        PADV2  & 30K & 30K & Verbs & 103 & / \\
        COCO-Tasks  & 40K & 64K & Phrases & 49 & 2.6 \\
        \midrule
        \multicolumn{6}{l}{\color{gray}Intention-Oriented Object Detection} \\
        RIO  & 40K & 130K & Template & 69 & 15.7 \\
        \midrule
        \multicolumn{6}{l}{\color{gray}Intention-Driven Visual Grounding} \\
        \rowcolor{mygray}
        \textbf{IntentionVG} & \textbf{100K} & \textbf{500K} & \textbf{Free-Form} & \textbf{1096} & 11.2 \\
    \specialrule{.1em}{.05em}{.05em}
    \end{tabular}
    \vspace{-2mm}
    \caption{Comparison with classic VG \cite{kazemzadeh2014referitgame,yu2016modeling,nagaraja2016modeling,liu2023gres}, referring video object segmentation \cite{seo2020urvos,kurita2023refego}, AD \cite{chuang2018learning,luo2021one,zhai2022one,sawatzky2019object} and RIO \cite{qu2023rio} datasets. 
    \# denotes the number, 
    where Intentions, Cats and Avg Len denote the intention expression types, object/affordance categories and average expression length.
    ``-'', ``/'' denote the intention and non-verb expressions are unavailable.}
    \label{tab:dataset_details}
    \vspace{-4mm}
\end{table}

\begin{figure}[htbp]
    \centering
    \includegraphics[width=0.49\textwidth]{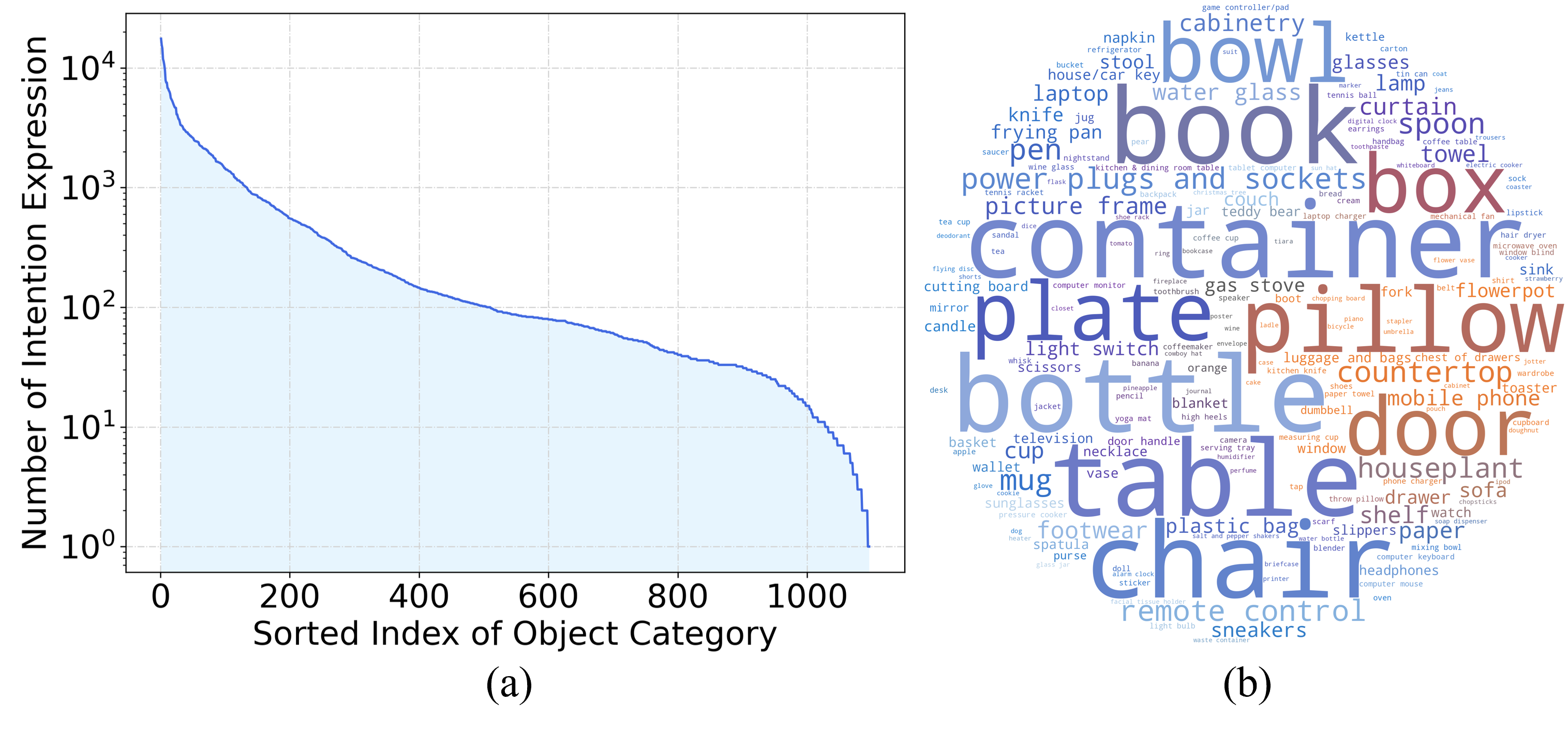}
    \vspace{-22pt}
    \caption{IntentionVG dataset statistics.
    (a) the number of referring expressions per object's category in the log scale. 
    (b) the word cloud highlights the head categories.
    }
    \label{fig_data_analysis1}
    \vspace{-10pt}
\end{figure}

\begin{figure}[htbp]
    \centering
    \includegraphics[width=0.49\textwidth]{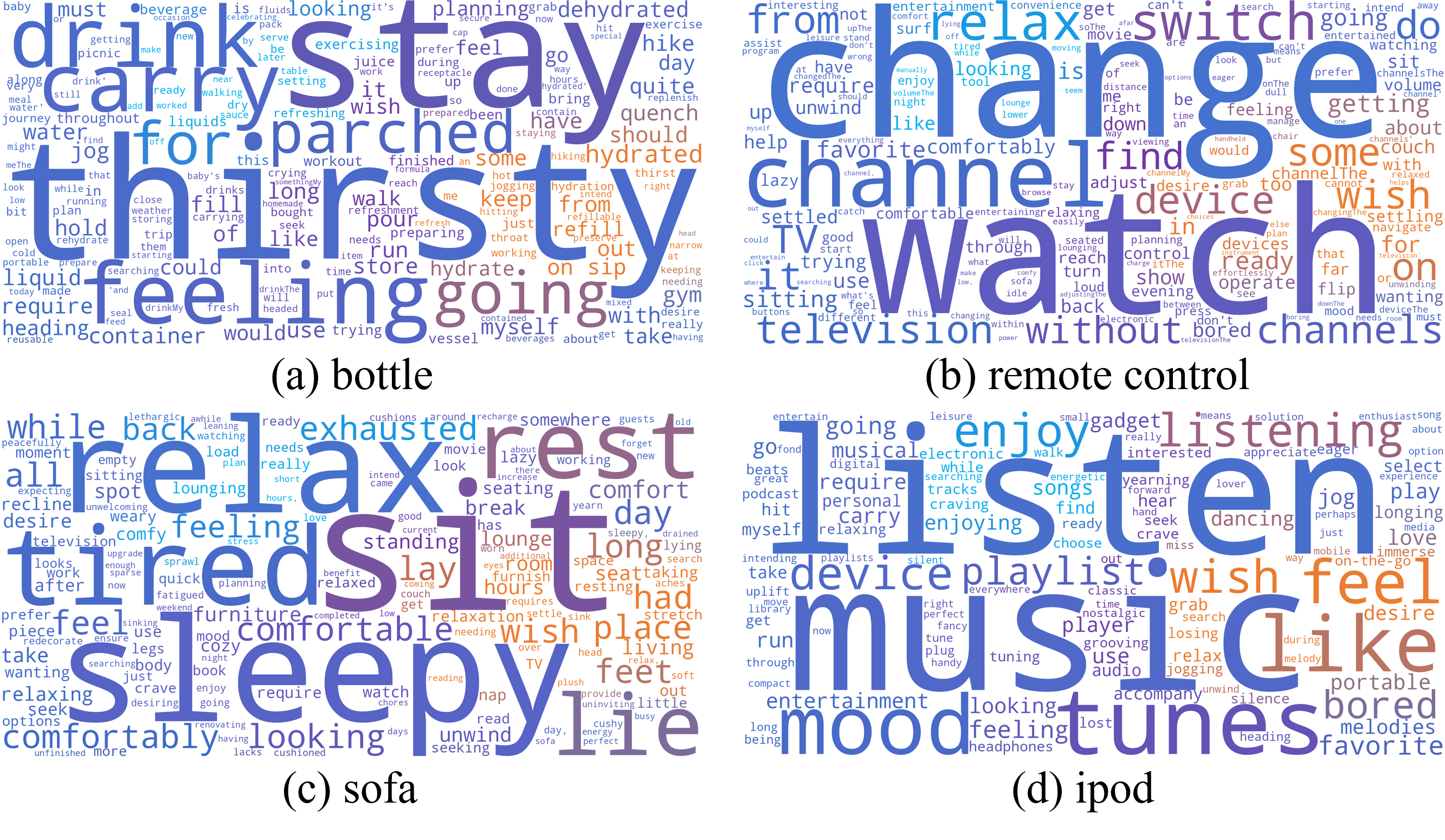}
    \vspace{-22pt}
    \caption{Word clouds of partial categories from our IntentionVG benchmark dataset.}
    \label{fig_data_analysis2}
    \vspace{-10pt}
\end{figure}

\begin{figure}[htbp]
    \centering
    \includegraphics[width=0.47\textwidth]{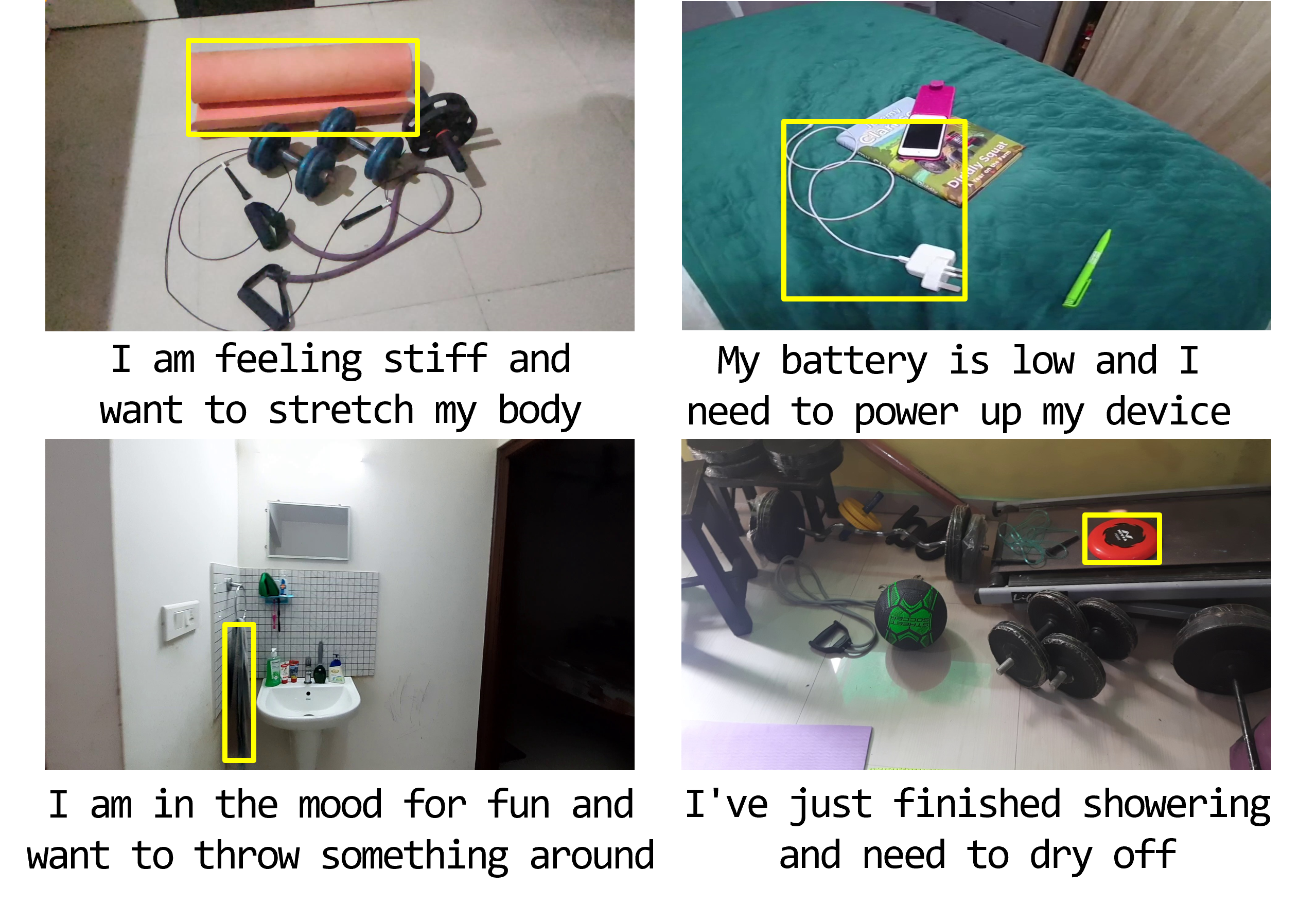}
    \vspace{-10pt}
    \caption{Visualizations of samples from our IntentionVG benchmark dataset.}
    \label{fig_IntentionVG_examples}
    \vspace{-15pt}
\end{figure}

As detailed in Table \ref{tab:dataset_details}, prevailing datasets, such as the widely-used benchmark RefCOCO \cite{yu2016modeling}, exhibit limitations in terms of small data scale and a scarcity of natural language expressions that reflect human intentions.
We compare the built IntentionVG dataset with existing benchmark (including the datasets for classic VG, AD, IOOD and our IVG tasks) to highlight the distinct and significant properties of our dataset, as outlined in Table \ref{tab:dataset_details}. 
Besides, the dataset statistics of IntentionVG are presented in Fig. \ref{fig_data_analysis1}, Fig. \ref{fig_data_analysis2} and Fig. \ref{fig_data_analysis3} in \textcolor{blue}{Appendix} Sec. \ref{subsec:more_visualizations}, illustrating its data diversity and potential for practical applications, while Fig. \ref{fig_IntentionVG_examples} and Fig. \ref{fig_more_IntentionVG_examples} in \textcolor{blue}{Appendix} Sec. \ref{subsec:more_visualizations} showcase the examples from our IntentionVG dataset.

\paragraph{Intention-Driven Descriptions.}
In comparison to the previous grounding counterparts, our newly built IntentionVG is the first visual grounding dataset covering free-form intention-oriented expressions for each object in the provided images. 
Compared with the affordance detection counterparts, our IntentionVG dataset transcends the closed-set affordance categories, providing informative and unique intention-driven language expressions for each bbox.

\vspace{-5pt}
\paragraph{Egocentric View \& Multi-Scene Perception.}
Unlike most existing grounding datasets, our IntentionVG is the first work to incorporate the egocentric perspective and multi-scene perception that are critically needed by multimodal agents in real-world scenarios. 
The grounding data within IntentionVG are annotated with scene categories, with its training and testing sets specifically designed to support model training and evaluation under multi-scene setting, which moves beyond the typical single-image input of classic VG and more closely aligns with application scenarios.

\vspace{-5pt}
\paragraph{Breakable Data Scales \& Object Categories.}
To the best of our knowledge, our IntentionVG stands as the largest-scale dataset within the grounding research community to date.
In terms of the number of images, object instances, and referential tokens, IntentionVG significantly outpaces the previous largest classic grounding dataset, RefCOCOg \cite{nagaraja2016modeling}, multiplying its scale by nearly 4, 9, and 12 times respectively.
Meanwhile, it encompasses intention-based expression counts that exceed the largest existing AD dataset COCO-Tasks \cite{sawatzky2019object} and IOOD dataset RIO \cite{qu2023rio} by 8, 4 times separately.  
Featuring 1096 object categories and nearly 500K expressions about user intentions, IntentionVG spans a broader spectrum of multimodal knowledge, marking a pivotal advancement in the pursuit of open-world intention understanding.

\vspace{-5pt}
\paragraph{More Complex and Free-Form References.}
Benefiting from the powerful GPT-4 \cite{achiam2023gpt} and our carefully crafted prompt template, the reference expressions of IntentionVG dataset are enriched with visual context to capture human intentions more effectively. 
Without sticking to a rigid template (\eg, [Something to ...] format in RIO \cite{qu2023rio}), our IntentionVG allows for the diverse intentions behind interacting with various target entities to be articulated and emphasized through flexible natural language expressions.

\subsection{Evaluation Metrics}
\label{subsec:metrics}
 
To advance the IVG task's applicability in practical scenarios, we have introduced two grounding task settings based on the quantity of images provided: single-scene and multi-scene grounding. 
We have also tailored evaluation metrics for each setting, enabling a thorough assessment of model performance across different contexts. 

\vspace{-5pt}
\paragraph{Single-Scene.} 
In this setting, the model's grounding capability is assessed with just one provided image. 
We use Precision@0.5 (P@0.5) as the metric to evaluate models' grounding performance. 
This measure reflects the model's ability to correctly identify the target object in alignment with the user's intention with its top one prediction. 
A prediction is considered correct if the Intersection over Union (IoU) between the predicted and the ground truth (GT) bbox exceeds a threshold of 0.5 (\ie, $threshold > 0.5$), indicating a significant overlap and, hence, an accurate localization result.

\vspace{-5pt}
\paragraph{Multi-Scene.} 
Since the intelligent agents need to move and search for the expected targets among different scenarios in practice, it is vital for embodied agents to accomplish our IVG task based on multi-scenario perception in the first-person view. 
To assess the accuracy of models in multi-scene perception and the subsequent VG, we utilize the metrics of Recall@1 (R@1) and Precision@0.5 (P@0.5). 
Recall@1 measures the ratio of cases accurately identified within the top-1 perception result to the overall count of cases in the test set, reflecting the model's precision in pinpointing the most intention-relevant image from a multitude of scenes.
Additionally, we introduce Precision@0.5 given Recall@1 correct cases (P@0.5|R@1) as an evaluation metric to gauge the grounding performance of models specifically for those cases correctly identified in the multi-scene perception phase. 
These employed metrics ensure a nuanced understanding of the models' effectiveness in accurately grounding objects across multiple scenes.

\section{Baseline Construction} 

To realize our IVG task based on multi-scene perception and egocentric viewpoint, we formulate two different kinds of baseline models below.
More illustrations of the baseline structures can be found in \textcolor{blue}{Appendix} Sec. \ref{subsec:baseline} for better understanding.

\subsection{Zero-shot Setting}
For zero-shot setting, it is intuitively to follow a two-step integrated approach to realize the multi-scene perception and VG for IVG task.
We initially employ EVA-CLIP \cite{sun2023eva} as perceiver to extract multi-scene representations and features of intention expressions, followed by feature similarity matching to select the scene that best matches the textual expression. 
Subsequently, the well matched single scene is fed to the grounding model with the corresponding intention expression to deduce the associated bounding box. 
The adopted grounding models can be categorized into two types: specialists and generalists. 
Specialist baselines are the SOTA methods designed for classic VG task, including MDETR \cite{kamath2021mdetr}, SeqTR \cite{zhu2022seqtr} and Polyformer \cite{liu2023polyformer}, which are trained on VG datasets.
Generalist baselines comprise models capable of handling various V-L tasks that are trained on large-scale visual question answering (VQA) and VG datasets, such as LLM-based method Shikra \cite{chen2023shikra} and Mini-GPTv2 \cite{chen2023minigpt}.
We also incorporate LLMs (\eg, GPT-4 \cite{achiam2023gpt}) as the optional interpreter which could translate intention expressions into explicit object descriptions, helping the following perceiver and grounding model better solve our IVG task. 

\begin{table*}[htbp]
    \small
    \setlength{\belowcaptionskip}{1.0pt}
    \begin{center}
    \setlength{\tabcolsep}{1.6mm}{
    \begin{tabular}{l|c|c|ccc}
    \specialrule{.1em}{.05em}{.05em}
        \multicolumn{1}{l|}{\multirow{3}{*}{Methods}} & \multicolumn{1}{c|}{\multirow{3}{*}{\parbox{1.6cm}{\centering Framework Type}}} & \multicolumn{4}{c}{\multirow{1}{*}{Intention-Driven Visual Grounding}} \\ 
        \cline{3-6}
        &  & \multicolumn{1}{c|}{Single-Scene} & \multicolumn{3}{c}{Multi-Scene} \\
        \cline{3-6}
         &   & P@0.5 & R@1 & P@0.5|R@1 & P@0.5 \\
        \specialrule{.1em}{.05em}{.05em}
        \multicolumn{1}{l|}{\textit{Zero-shot Setting}} & \multicolumn{4}{l}{} \\
        \midrule
        Perceiver + MDETR \cite{kamath2021mdetr} & Integrated  & 14.23 & 54.00 & 16.48 & 8.90    \\
        Perceiver + SeqTR \cite{zhu2022seqtr} & Integrated  & 9.38 & 54.00 & 9.38 & 5.07 \\
        Perceiver + Polyformer \cite{liu2023polyformer} & Integrated  & 16.50 & 54.00 & 18.64  & 10.07 \\
        Perceiver + Grounding DINO \cite{liu2023grounding} & Integrated  & 14.51	& 54.00	& 16.99	& 9.18 \\
        \midrule
        Perceiver + OFA  \cite{wang2022ofa}  & Integrated  & 18.57 & 54.00 & 18.57 & 10.03 \\
        Perceiver + Shikra~\cite{chen2023shikra} & Integrated & 22.45 & 54.00 & 24.44 & 13.20 \\
        Perceiver + Ferret~\cite{you2023ferret} & Integrated &  21.80 & 54.00 & 24.19 & 13.06 \\
        Perceiver + LISA~\cite{lai2023lisa} & Integrated & 15.32 & 54.00 & 17.73 & 11.08 \\ 
        Perceiver + Qwen-VL~\cite{bai2023qwen} & Integrated & 20.72 & 54.00 & 22.71 & 12.27 \\
        Perceiver + MiniGPT-v2~\cite{chen2023minigpt} & Integrated & 13.04 & 54.00 & 14.51 & 7.84 \\
        \midrule
        Interpreter + Perceiver + Shikra~\cite{chen2023shikra} & Integrated & {37.55} & {62.63} & {42.35} & {26.45} \\
        Interpreter + Perceiver + MiniGPT-v2~\cite{chen2023minigpt} & Integrated & 42.57 & 62.63 & 45.80 & 28.68 \\
        \midrule
        \multicolumn{1}{l|}{\textit{Fine-tuning Setting}} & \multicolumn{4}{l}{} \\
        \midrule
        Perceiver + SeqTR~\cite{zhu2022seqtr} & Integrated & 36.69 & 62.63 & 42.99 & 26.75 \\ 
        Perceiver + Shikra~\cite{chen2023shikra} & Integrated & 47.19 & 62.22 & 50.43 & 31.38 \\ 
        Perceiver + MiniGPT-v2~\cite{chen2023minigpt}  & Integrated & 44.18  & 62.22 & 49.92 & 31.06  \\
        \midrule
        \rowcolor{mygray} Qwen-VL~\cite{bai2023qwen} & End-to-End & \textbf{50.58} &  \textbf{74.13} & \textbf{53.02} & \textbf{39.30} \\
        \specialrule{.1em}{.05em}{.05em}
    \end{tabular}
    \vspace{-2mm}
    \caption{Comparisons with the classic VG SOTA approaches on our ItentionVG testing set. 
    ``Perceiver" and ``Interpreter" separately denote the EVA-CLIP model and GPT-4 employed for multi-scene perception and user intention understanding.
    Besides, ``Integrated" and ``End-to-End" respectively refer to the baseline is a grounding model combined with Perceiver or Interpreter and the single grounding model as an end-to-end structure.
    }
    \label{tab:sota}}
    \end{center}
    \vspace{-20pt}
\end{table*}

\subsection{Fine-tuning Setting}
For fine-tuned baselines, we incorporate both integrated and end-to-end models.
For the two-step integrated baselines, we first utilize contrastive learning loss to fine-tune EVA-CLIP \cite{sun2023eva} with the annotations of scene categories, enhancing its capability of discerning the similarity between scenes and intention expressions. 
Then we perform fine-tuning on generalist grounding models using their respective training prompts for grounding task, followed by putting together the fine-tuned scene perceiver and grounding models.

The end-to-end baseline model is built upon generalist Qwen-VL \cite{bai2023qwen}. 
Since only Qwen-VL possesses the capability to accommodate multiple images as inputs, we introduce a well-designed prompt that facilitates the simultaneous input of multiple scenes and intention expressions and build the end-to-end baseline upon Qwen-VL. 
Through instruction tuning under cross entropy loss, our fine-tuned Qwen-VL can concurrently conduct scene perception and grounding.

To be noticed, since target object matching the intent description may appear in multiple scenes simultaneously, to ensure that only the positive scene contains the corresponding object, we impose a strong constraint on the input scenes to force that only one of them exists the object.
Besides, two hyper-parameters including input scene number N and multi-scene occurrence rate $\alpha$ are introduced during fine-tuning.
A higher number N of input scenes implies that the baseline models need to identify the most relevant scene and target object from a larger set of images during fine-tuning, thus increasing the difficulty of the training objective.
Besides, a higher $\alpha$ value means that the baseline models are more likely to encounter multi-scene input samples during fine-tuning, as opposed to the classic VG's typical single-image input. 
$\alpha$=0 or 1.0 represents extreme circumstances during fine-tuning, where the model is exclusively fed either single-scene or multi-scene input samples.

\section{Experiments} 
\label{sec:expe}

To evaluate the effectiveness and the designing rationale of our data and baseline models, comprehensive experiments are conducted on 
our built IntentionVG dataset for the new IVG task.

\subsection{Implementation Details}

Our work is implemented based on Pytorch \cite{paszke2019pytorch} and trained with 8 NVIDIA A800 GPUs. 
The original weights of all the adopted baselines are inherited for the subsequent fine-tuning and evaluations under zero-shot setting. 
For baseline constructions under fine-tuning setting, we either directly fine-tune the end-to-end baseline (\ie, Qwen-VL \cite{bai2023qwen}) or put together the separately fine-tuned intention interpreter, multi-scene perceiver and grounding models for the integrated baseline construction.
Taking EVA-CLIP \cite{sun2023eva} as scene perceiver, we introduce the annotations of bounding boxes and scene categories to respectively tune the grounding models and EVA-CLIP.
Training details about fine-tuning are presented in Table \ref{table_implementation_details} in \textcolor{blue}{Appendix} Sec. \ref{subsec:implementation_details}.

\subsection{Main Results and Analysis}
\label{subsec:main_results}

To quantitatively evaluate the intention-oriented grounding performance of all the constructed baseline models for our new IVG task, we conduct experimental comparison on the newly built IntentionVG testing set.
As illustrated in Table \ref{tab:sota}, the baseline models can be categorized into two settings based on the evaluation manner: zero-shot and fine-tuning, and into two types based on the framework structure: integrated and end-to-end.
For a fair comparison, we re-implement these SOTA methods and report their performance on our IntentionVG testing set.
It is evident from Table \ref{tab:sota} that the zero-shot setting baselines, having not been exposed to our intention-driven grounding data, struggle to comprehend user intentions and identify corresponding targets, resulting in generally lower performance compared to the fine-tuned baselines.
With EVA-CLIP serving as the multi-scene perceiver, neither specialists nor generalists classic VG SOTA methods can effectively address our IVG task directly, particularly in the multi-scene setting.
In contrast, baselines under the fine-tuning setting, both integrated and end-to-end types, have witnessed significant performance improvements on our IVG task with the support of IntentionVG data. 
Additionally, the introduction of the LLM-based intention interpreter has substantially enhanced the performance of zero-shot baselines on the IVG task, underscoring the critical importance of genuinely comprehending user intentions for accurately accomplishing the intention-driven grounding task.
Due to the significantly higher difficulty of our IVG task's multi-scene setting compared to single-scene counterpart, the multi-scene accuracy value is correspondingly lower than the other one. 
This further emphasizes the vital importance of researching intention-oriented visual grounding where previous SOTA methods have fallen short.

\begin{figure}[thbp]
    \centering
    \includegraphics[width=0.49\textwidth]{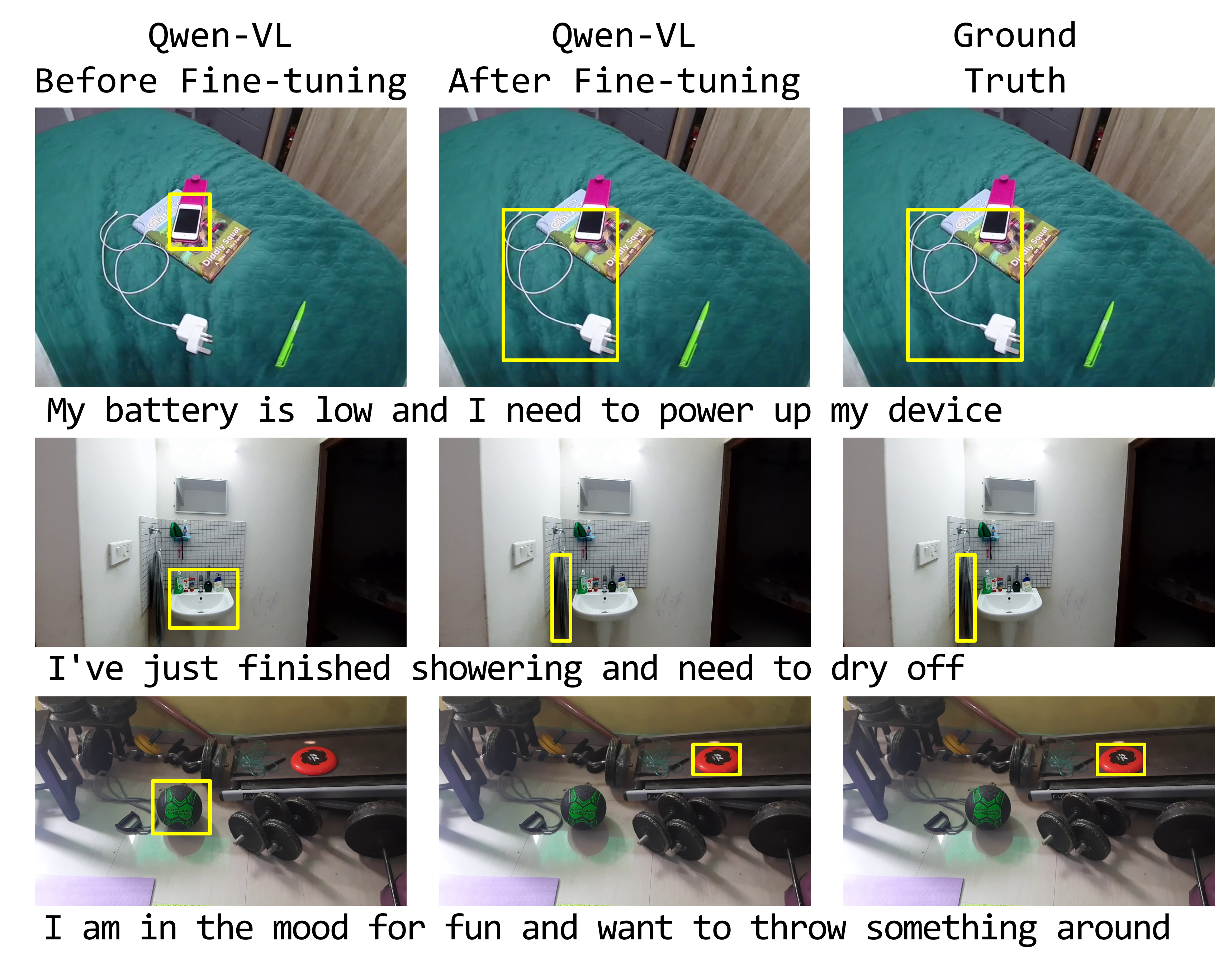}
    \vspace{-22pt}
    \caption{The visual comparison of baseline's predictions before and after fine-tuned on  IntentionVG dataset.}
    \label{fig_pre-post_finetune}
    \vspace{-15pt}
\end{figure}

\begin{figure}[thbp]
    \centering
    \includegraphics[width=0.49\textwidth]{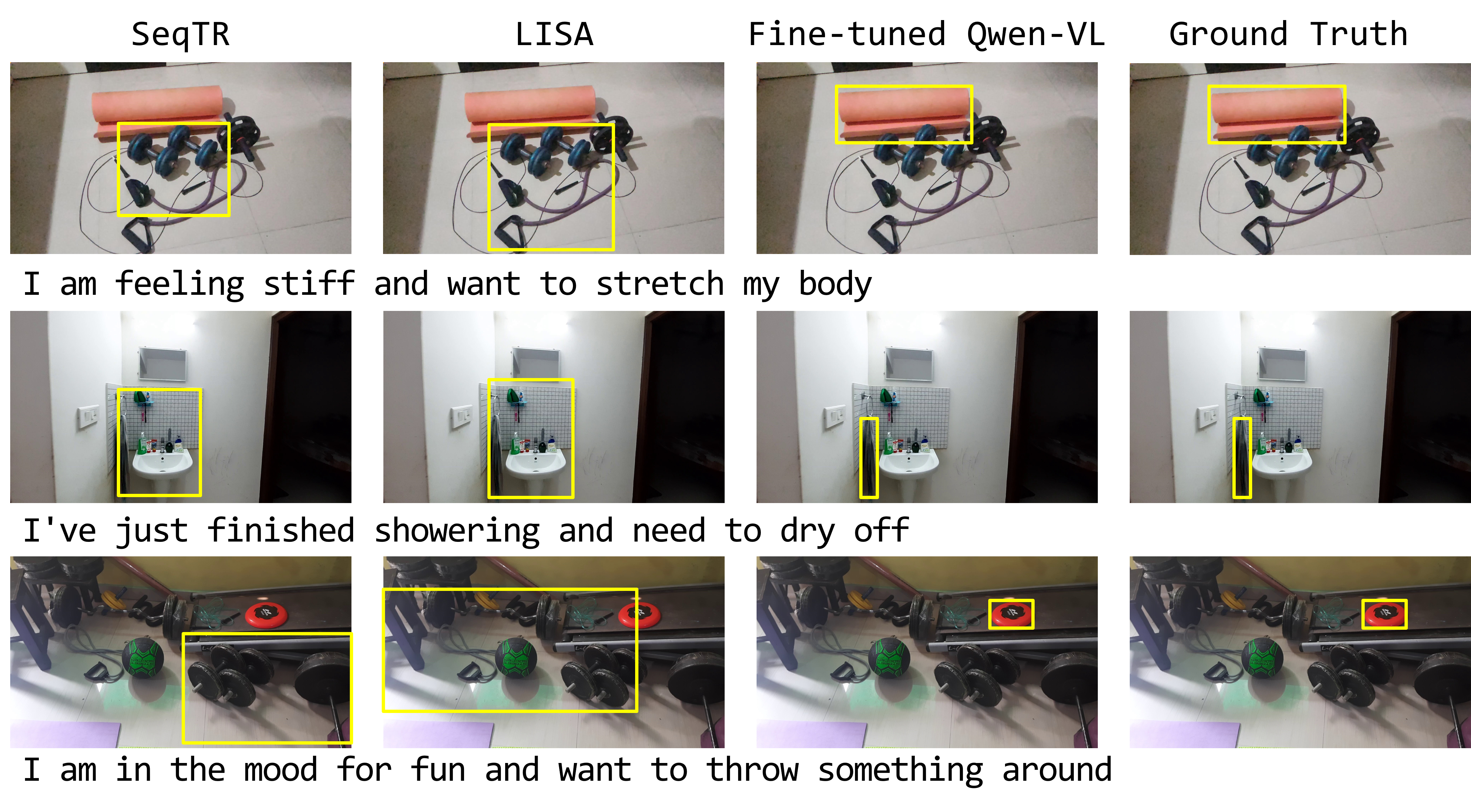}
    \vspace{-22pt}
    \caption{The visual comparison of grounding results between different baselines on IntentionVG dataset.}
    \label{fig_qualitative_comparison}
    \vspace{-15pt}
\end{figure}

For qualitative analysis, we further present the qualitative results of taking Qwen-VL as baseline model before and after being fine-tuned on our IntentionVG dataset, the prediction comparisons between several different baseline models, which can be respectively found in Fig. \ref{fig_pre-post_finetune} and Fig. \ref{fig_qualitative_comparison}. 
It's clear to see in Fig. \ref{fig_pre-post_finetune} that before fine-tuned on our IntentionVG dataset, the baseline model could not accurately understand user intentions and locate the targets corresponding to the given intention descriptions. However, after fine-tuning on our IntentionVG data, the models significantly improve their capability in the IVG task based on understanding user intentions.
Moreover, as shown in Fig. \ref{fig_qualitative_comparison}, both the LLM-based SOTA REC method LISA \cite{lai2023lisa} and the non-LLM-based SOTA REC method SeqTR \cite{zhu2022seqtr} struggle to accurately locate the targets matching user intentions under the zero-shot setting. However, models (\ie, Qwen-VL~\cite{bai2023qwen}) fine-tuned with data from our IntentionVG dataset can accurately locate the corresponding targets, achieving results with high consistency with the real labels.

\subsection{Ablation Study} 

To justify the efficacy of our IntentionVG's data, we conduct extensive ablation experiments on IntentionVG testing set. 
As illustrated in \ref{subsec:metrics}, the tables below involve both the traditional single-scene and harder multi-scene settings.
For all ablations, 
Qwen-VL~\cite{bai2023qwen} stays employed as our baseline and the multi-scene input is adopted as 5 images for fair evaluation. 
Except for the 1st ablation, 10\% of our IntentionVG data is used to fine-tune the baseline for ablations. 
Noticeably, more ablations can be found in \textcolor{blue}{Appendix} Sec. \ref{subsec:more_ablation}.

\vspace{-6pt}
\paragraph{Effect of Data Scale.}
First, we investigate the effect of different percentages of introduced training samples in our IntentionVG dataset. 
The results are shown in Table \ref{tab:data_scale}. 
It is clear in Table \ref{tab:data_scale} that model performance under both settings for our intention-oriented grounding task is consistently improved with more and more training samples, validating the efficacy and high quality property of our collected data. 
\begin{table}[thbp]
    \small
    \setlength{\belowcaptionskip}{1.0pt}
    \begin{center}
     \setlength{\tabcolsep}{1.2mm}{\begin{tabular}{c|c|ccc}
      \specialrule{.1em}{.05em}{.05em} 
      \multirow{2}{*}{Ratios} & \multirow{1}{*}{Single-Scene} &  \multicolumn{3}{c}{Multi-Scene} \\
    \cline{2-5}
          & P@0.5  & R@1 & P@0.5|R@1 & P@0.5  \\
    \specialrule{.1em}{.1em}{.1em}
        0\% & 20.72 & 0.00 & 0.00 & 0.00 \\ 
        10\% & 46.01 & 70.64 & 48.96 & 34.58 \\ 
        25\% & 47.63 & 73.46 & 50.67 & 37.22  \\ 
        50\% & 48.75 & 73.82 & 51.66 & 38.14  \\
    \rowcolor{mygray} 100\% & \textbf{50.58} &  \textbf{74.13} & \textbf{53.02} & \textbf{39.30} \\
    \specialrule{.1em}{.05em}{.05em}
    \end{tabular}
     \vspace{-2mm}
    \caption{Ablation study on effect of data scale.}
    \vspace{-6mm}
    \label{tab:data_scale}}
    \end{center}
\end{table}
Since the original Qwen-VL can not handle the multi-scene grounding task at all, when ratios=0 (\ie, without employing our intention-based data) the corresponding performance is much poor.
As the ratios of employed training samples continually rise, there's no sign of diminishing performance gains, suggesting that our dataset has great potential to help multimodal large language models better understand human intentions with consistently scaled up training data.

\vspace{-1mm}
\begin{table}[htbp]
    \small
    \setlength{\belowcaptionskip}{1.0pt}
    \begin{center}
     \setlength{\tabcolsep}{1.2mm}{\begin{tabular}{c|c|ccc}
      \specialrule{.1em}{.05em}{.05em} 
      \multirow{2}{*}{Scenes (N)} & \multirow{1}{*}{Single-Scene} &  \multicolumn{3}{c}{Multi-Scene} \\
      \cline{2-5}
          & P@0.5  & R@1 & P@0.5|R@1 & P@0.5  \\
    \specialrule{.1em}{.1em}{.1em}
        1 & 45.51 & 19.37 & 1.34 & 0.26 \\ 
        3 & 45.35 & 51.17 & 37.52 & 19.20 \\ 
        \rowcolor{mygray} 5 & 46.01 & \textbf{70.64} & \textbf{48.96} & \textbf{34.58}  \\ 
        8  &  \textbf{46.14} & 70.32 & 40.90 & 28.76  \\
        10 & 45.99 & 60.97 & 32.66 & 19.91  \\
    \specialrule{.1em}{.05em}{.05em}
    \end{tabular}
    \vspace{-2mm}
    \caption{Ablation study on effect of scene number N.}
    \vspace{-8mm}
    \label{tab:scene_number}}
    \end{center}
\end{table}

\paragraph{Effect of Scene Number.}
Furthermore, we explore the impact of varying the input scene number during training. 
Scene number N implies that the baseline model need to identify the most relevant scene and target object from a set of given images during fine-tuning.
As shown in Table \ref{tab:scene_number}, the model achieves optimal performance with an intermediate scene number of 5 under both settings. 
As scene number gradually increases from 1 to 10, the impact on the model's single-scene performance remains minimal. 
However, under multi-scene setting, the model's ability to perform multi-scene perception and subsequent grounding initially improves and then diminishes.
We believe this pattern occurs because, with few input scenes at the start, the model's learned capability in perception and grounding during tuning phase is weak. 
As the number of scenes increases, the model's related abilities enhance. 
Yet, beyond the sweet spot N=5, the model becomes overwhelmed due to the excessive number of input scenes, leading to confusion and an inability to learn the effective representations to accurately complete our IVG task.

\vspace{-1mm}
\begin{table}[htbp]
    \small
    \setlength{\belowcaptionskip}{1.0pt}
    \begin{center}
     \setlength{\tabcolsep}{1.2mm}{\begin{tabular}{c|c|ccc}
      \specialrule{.1em}{.05em}{.05em} 
      \multirow{2}{*}{Rates ($\alpha$)} & \multirow{1}{*}{Single-Scene} &  \multicolumn{3}{c}{Multi-Scene} \\
    \cline{2-5}
    & P@0.5  & R@1 & P@0.5|R@1 & P@0.5  \\
    \specialrule{.1em}{.1em}{.1em}
        0 & 45.70  & 19.71  & 1.48  & 0.29  \\ 
        0.25 & 44.57 & 46.41 & 38.81 & 18.02 \\ 
        0.5 & 45.54 & 70.61 & 48.58 & 34.31 \\  
        0.75 & 45.92 & 68.84 & 48.43 & 33.34 \\  
        \rowcolor{mygray} 0.9 & \textbf{46.01} & 70.64 & \textbf{48.96} & 34.58  \\ 
        1.0 & 42.77 & \textbf{71.68} & 48.45 & \textbf{34.73} \\ 
    \specialrule{.1em}{.05em}{.05em}
    \end{tabular}
     \vspace{-2mm}
    \caption{Ablation study on the effect of multi-scene Occurrence rate during fine-tuning.}
    \vspace{-8mm}
    \label{tab:multiscene_rate}}
    \end{center}
\end{table}

\paragraph{Effect of Multi-Scene Occurrence Rate.}
Additionally, we delve into the impact of the multi-scene occurrence rate, denoted as $\alpha$. 
$\alpha$ means that the possibility of baseline models to encounter multi-scene input samples during fine-tuning. 
It is evident from Table \ref{tab:multiscene_rate} that when $\alpha$=0, the model performs poorly on our multi-scene IVG task due to the lack of multi-scene samples during tuning. 
A larger $\alpha$ helps the model achieve better grounding performance based on multi-scene perception. 
However, an excessively high $\alpha$ value, specifically $\alpha$=1.0, significantly diminishes the performance in intention-driven grounding tasks with single-image inputs. 
Therefore, 
$\alpha$=0.9 is set as our default setting.

\vspace{-1mm}
\begin{table}[htbp]
    \small
    \setlength{\belowcaptionskip}{1.0pt}
    \begin{center}
     \setlength{\tabcolsep}{0.6mm}{\begin{tabular}{cc|c|ccc}
      \specialrule{.1em}{.05em}{.05em} 
      \multicolumn{2}{c|}{Supervision Types} & \multirow{1}{*}{Single-Scene} &  \multicolumn{3}{c}{Multi-Scene} \\
    \cline{3-6}
       Scene & Object & P@0.5  & R@1 & P@0.5|R@1 & P@0.5  \\
    \specialrule{.1em}{.1em}{.1em}
        \fmark & \fmark & 45.89 & 19.84  & 1.63  & 0.32  \\
        \cmark & \fmark  & 44.76 & 59.46 & 45.65  & 27.15\\ 
        \fmark & \cmark & 45.70  & 19.37  & 1.34  & 0.26 \\
        \rowcolor{mygray} \cmark & \cmark & \textbf{46.01} & \textbf{70.64} & \textbf{48.96} & \textbf{34.58}\\ 
    \specialrule{.1em}{.05em}{.05em}
    \end{tabular}
    \vspace{-2mm}
    \caption{Ablation study on effect of GT formality.}
    \vspace{-8mm}
    \label{tab:response_formality}}
    \end{center}
\end{table}

\paragraph{Effect of Supervision Formality.}
At last, we exploit the effect of the GT formality for supervision during training.
The supervision signal includes four different settings, which are only grounding bbox, bbox + scene category, bbox + object tag and bbox + scene category + object tag (the 1st to 4th row in Table \ref{tab:response_formality}).
As shown in Table \ref{tab:response_formality}, 
with the bbox, scene category, and object tag serving as the most comprehensive supervision targets, the model gains access to the fullest extent of information and establishes explicit associations between inputs and GTs, thereby achieving best performance. 
The inclusion of scene category and object tag as part of the supervision signal respectively enhances the model's multi-scene perception and open-domain object recognition capabilities. 
Thus, removing either of these two parts leads to performance decline on our IVG task. 
From the 1st and 3rd rows we can observe that introducing the object tag as an additional part of GT on top of bbox does not result in performance improvement. 
We believe this is because, without the guidance of the scene category, the model becomes very confused during fine-tuning and fails to learn a clear mapping from multi-scene inputs to the target scene and subsequent grounding results. 
Consequently, the performance of these two rows remains poor on the multi-scene grounding setting.

\section{Conclusion and Broader Impact} 
\label{sec:conclusion}

In this paper, 
we move beyond previous works that focused solely on literal description based grounding and take a step further to intention-driven V-L understanding. 
Specifically, by considering that the practical agents need to move and search for expected targets among different scenarios, we put forward a new IVG task. 
The IVG task requires agents to interpret user intentions and locate specific targets based on egocentric view and multi-scene perception. 
To enable existing models to better accomplish our IVG task and assess their capabilities on it, we build the first also the largest-scale intention-driven grounding dataset termed IntentionVG with free-form expressions and develop a series of baseline to accomplish the IVG task. 

Comprehensive experiments conducted on our Intention dataset demonstrate that most previous methods struggle to directly understand users' non-literal intent expressions and locate the expected targets. 
With the aid of our data, there is a significant enhancement in the ability to comprehend intentions for IVG task, but there remains substantial room for improvement, warranting further investigations by the research community.
Aspiring to foster future research into intention-oriented grounding and inspire new research in this direction, we plan to release our newly built IntentionVG dataset and baseline models to the community.

\section*{Limitations}

While this work significantly advances the classic visual grounding task towards user intention-based grounding that aligns more closely with real-world applications, it is not without limitations, which leaves opportunities and room for future research. 
One potential limitation of this work is the scale of the IntentionVG benchmark dataset and the capacity of the developed baseline models could both be expanded and boosted to enhance performance on our proposed intention-driven visual grounding task. 
Additionally, this research primarily focuses on the referring expression comprehension and generating bounding boxes as grounding output results, indicating that while the built baseline frameworks showcase new intention-driven grounding capabilities, they currently can not generate dense segmentation masks for the target objects. 
However, integrating with visual foundation models, such as the Segment Anything Model \cite{kirillov2023segment} and its variations, could equip our framework with the ability to produce the dense referring segmentation masks, effectively addressing this shortcoming. 
This possibility opens a new avenue for research, aiming to create a more versatile and powerful framework that leverages both large language models and visual foundation models to interact with user-provided intention-driven natural language expressions and visual perception inputs.


\section*{Acknowledgements}
We thank all the insightful reviewers for the helpful suggestions. 
This work was supported by the National Science and Technology Major Project (No.2022ZD0118801), National Natural Science Foundation of China (U21B2043, 62206279).

\clearpage

\bibliography{main}

\clearpage
\appendix
\section{Appendix}
\label{sec:appendix}

In this appendix, we provide the following items:
\begin{itemize}[noitemsep,leftmargin=*]
    \item (Sec. \textcolor{red}{1}) More implementation details on the built IntentionVG dataset for IVG task. 
    \item (Sec. \textcolor{red}{2}) More ablation studies on our newly built IntentionVG dataset for our IVG task. 
    \item (Sec. \textcolor{red}{3}) More illustrations about the constructed baselines for our IVG task. 
    \item (Sec. \textcolor{red}{4}) More visualizations of the dataset statistics and samples from our IntentionVG dataset. 
\end{itemize}

\subsection{Implementation Details}
\label{subsec:implementation_details}

\begin{table}[htbp]
\footnotesize
\addtolength{\tabcolsep}{2.6pt}
    \begin{tabular}{c|cc}
        \specialrule{.1em}{.05em}{.05em}
        \multirow{2}{*}{Configuration} & \multicolumn{2}{c}{Fine-tuning} \\
        \cline{2-3}
         &  Perceiver &  Grounding Model  \\
        \specialrule{.1em}{.05em}{.05em}
        \centering
        Optimizer & LAMB & AdamW \\
        Base Lr &0.0005 &  0.00001 \\
        Weight Decay &0.05  & 0.1  \\
        Batch Size  & 2048 & 32 \\
        Lr Decay Schedule & cosine &  cosine\\
        Training Epochs  & 25 & 1 \\     
        \specialrule{.1em}{.05em}{.05em}
    \end{tabular}
\vspace{-5pt}
\caption{Training settings on our IntentionVG dataset.}
\label{table_implementation_details}
\end{table}

The specific training hyper-parameter configurations for fine-tuning all the baseline models on our newly constructed IntentionVG dataset can be found in Table \ref{table_implementation_details}.

\subsection{More Ablation Studies}
\label{subsec:more_ablation}

\vspace{-1mm}
\begin{table}[htbp]
    \small
    \setlength{\belowcaptionskip}{1.0pt}
    \begin{center}
     \setlength{\tabcolsep}{0.3mm}{\begin{tabular}{c|c|ccc}
      \specialrule{.1em}{.05em}{.05em} 
      \multirow{2}{*}{Expression Forms} & \multirow{1}{*}{Single-Scene} &  \multicolumn{3}{c}{Multi-Scene} \\
    \cline{2-5}
        & P@0.5  & R@1 & P@0.5|R@1 & P@0.5  \\
        \specialrule{.1em}{.1em}{.1em}
        Object Tag & 14.39 & 30.71 & 21.57 & 6.62 \\ 
        \midrule
        {[Something to...]} & 44.70 & 66.24 & 46.74 & 30.96 \\ 
        {[I want to...]} & 43.80 & 47.24 & 40.10 & 18.95 \\ 
    \rowcolor{mygray} Free Form & \textbf{46.01} & \textbf{70.64} & \textbf{48.96} & \textbf{34.58} \\
    \specialrule{.1em}{.05em}{.05em}
    \end{tabular}
     \vspace{-2mm}
    \caption{Ablation study on the effect of expression form in our IntentionVG dataset.}
    \vspace{-8mm}
    \label{tab:expression_form}}
    \end{center}
\end{table}

\paragraph{Effect of Expression Form.}
We also probe into the effect of our grounding data's expression form. 
As presented in Table \ref{tab:expression_form}, the baseline model achieves the best result under both settings, by learning associations between more freely enriched expressions and target objects. 
During the fine-tuning phase, relying solely on the target object's tag as textual description can lead to an excessive dependence on direct target description, significantly reducing accuracy on our IVG task.
Besides, either changing the expression form of our IntentionVG data from free formality to the fixed counterpart (\ie, following the fixed template [Something to...] or [I want to...] to describe affordance or intention) will result in an considerable accuracy decrease across the two settings, which greatly impedes the potential for practical applications.

\subsection{Baseline Structures}
\label{subsec:baseline}

\begin{figure}[htbp]
    \centering
    \includegraphics[width=0.49\textwidth]{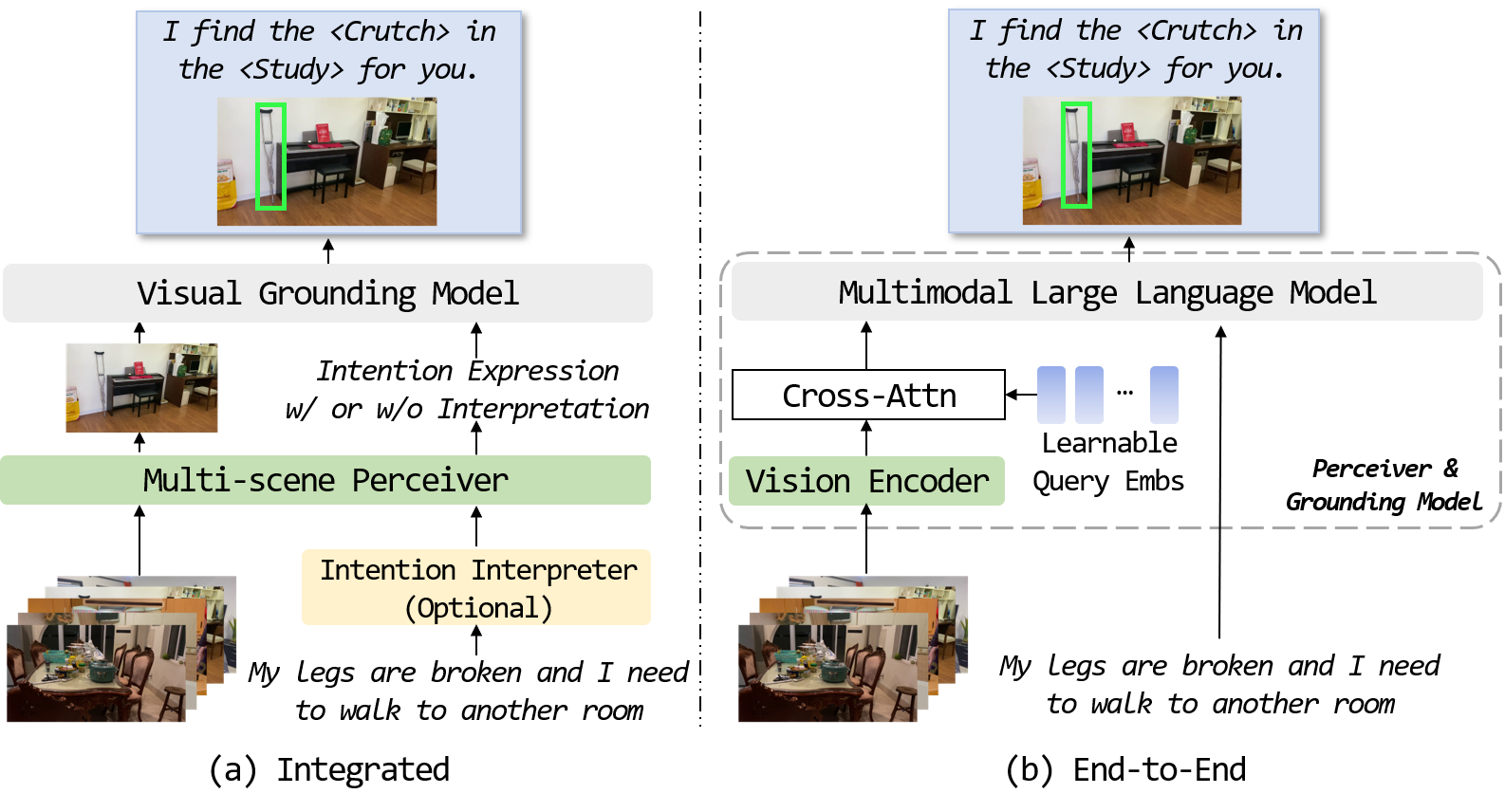}
    \vspace{-18pt}
    \caption{
    The illustration about the overall pipeline of our built baseline models with both integrated and end-to-end structures for the proposed IVG task.
    }
    \label{fig_baseline}
    \vspace{-10pt}
\end{figure}

We have also provided more illustrations about the constructed baselines of two types (\ie, integrated and end-to-end) in Fig. \ref{fig_baseline}.

\vspace{-1mm}

\subsection{More Visualization Results}
\label{subsec:more_visualizations}

\begin{figure*}[htbp]
    \centering
    \includegraphics[width=0.92\textwidth]{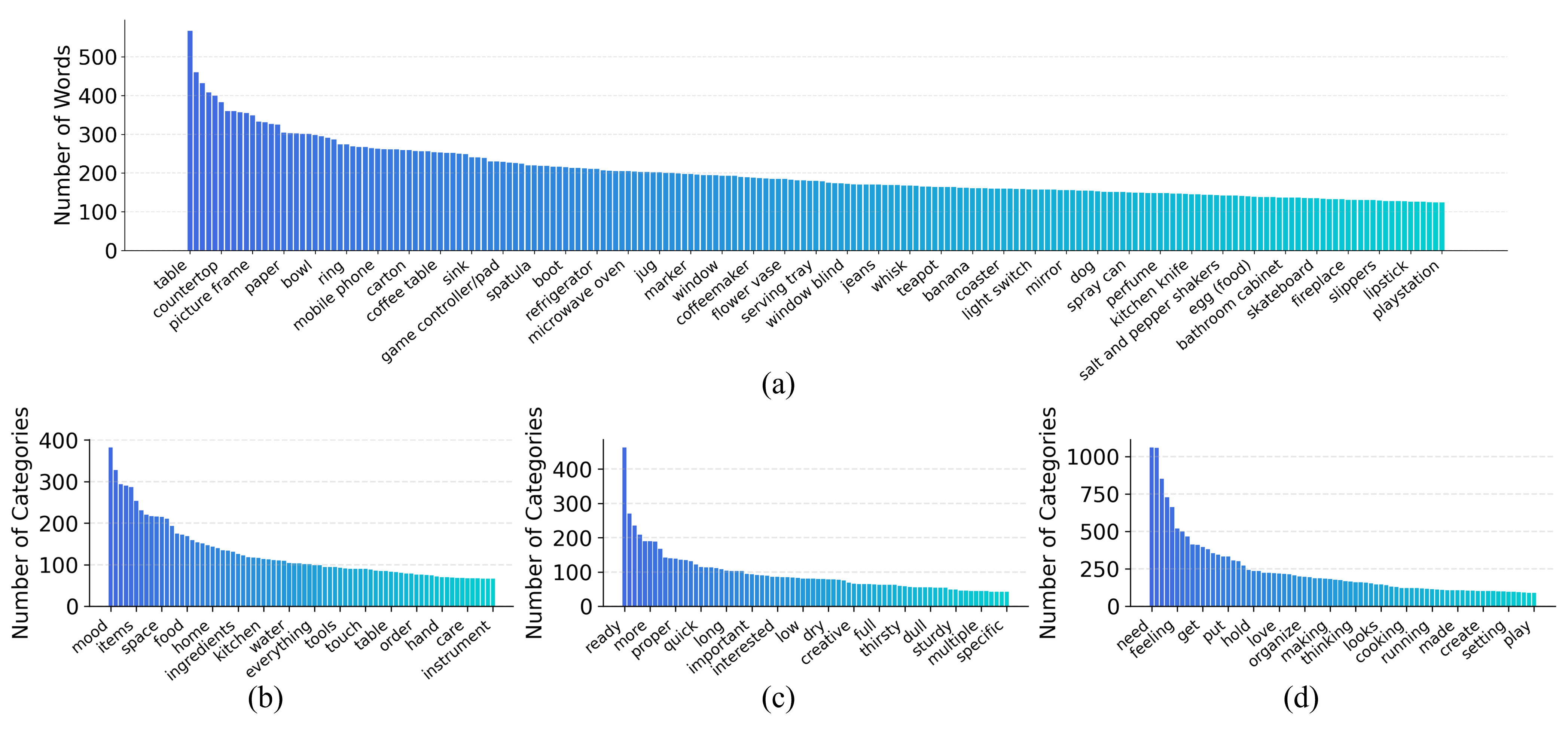}
    \vspace{-5pt}
    \caption{Our IntentionVG dataset statistics. 
    (a) shows the statistics of the word diversity in intention descriptions for each category, 
    and (b), (c), (d) separately present the occurrence frequency of a noun, adjective, verb in different categories of intention description. 
    The horizontal coordinates for (a), (b), (c) and (d) are respectively the examples of the specific categories, nouns, adjectives and verbs with the ranked top 200, 75, 75 and 75 highest vertical values.}
    \label{fig_data_analysis3}
    \vspace{-20pt}
\end{figure*}

\noindent \textbf{More IntentionVG Dataset Statistics.} 
More data statistics information about our newly built IntentionVG dataset are presented in Fig. \ref{fig_data_analysis3}. 

\begin{figure*}[htbp]
    \centering
    \includegraphics[width=0.92\textwidth]{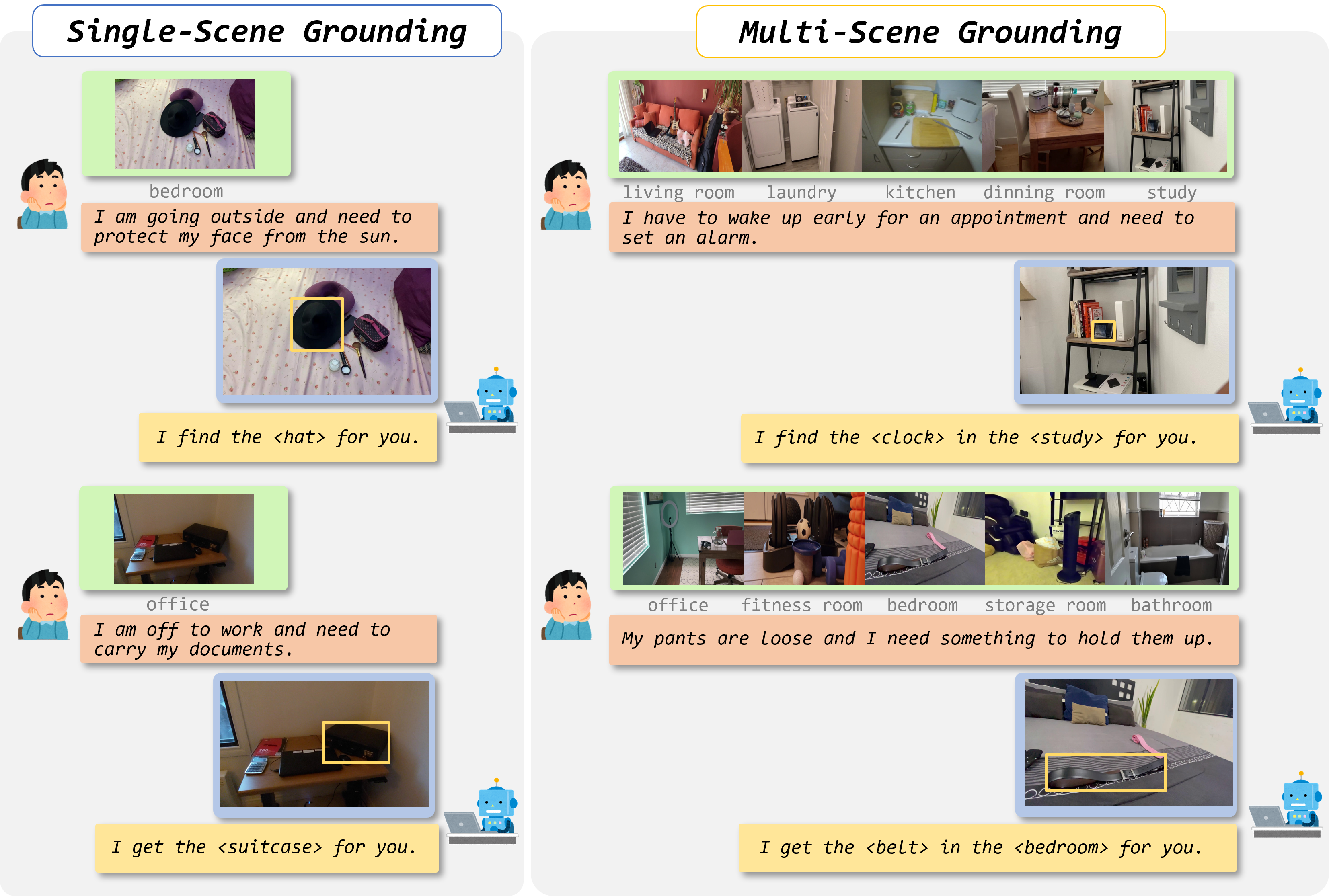}
    \caption{
    Visualizations of samples from our IntentionVG benchmark dataset. 
    }
    \label{fig_more_IntentionVG_examples}
\end{figure*}

\noindent \textbf{Samples of IntentionVG Dataset.} 
A few examples in our newly built IntentionVG dataset for intention-driven visual grounding task are presented in Fig. \ref{fig_more_IntentionVG_examples}. 

\end{document}